# Generalizing Machine Learning Evaluation through the Integration of Shannon Entropy and Rough Set Theory

Olga Cherednichenko[1], Dmytro Chernyshov[2], Dmytro Sytnikov[2] and Polina Sytnikova[2]

[1] University of Lyon 2, 5 avenue Mendès, Lyon, 69676, France
[2] National University of RadioElectronics, Nauky ave. 14, Kharkiv, 61166, Ukraine

**Abstract**

This research paper delves into the innovative integration of Shannon entropy and rough set theory, presenting a novel approach to generalize the evaluation approach in machine learning. The conventional application of entropy, primarily focused on information uncertainty, is extended through its combination with rough set theory to offer a deeper insight into data's intrinsic structure and the interpretability of machine learning models. We introduce a comprehensive framework that synergizes the granularity of rough set theory with the uncertainty quantification of Shannon entropy, applied across a spectrum of machine learning algorithms. Our methodology is rigorously tested on various datasets, showcasing its capability to not only assess predictive performance but also to illuminate the underlying data complexity and model robustness. The results underscore the utility of this integrated approach in enhancing the evaluation landscape of machine learning, offering a multi-faceted perspective that balances accuracy with a profound understanding of data attributes and model dynamics. This paper contributes a groundbreaking perspective to machine learning evaluation, proposing a method that encapsulates a holistic view of model performance, thereby facilitating more informed decision-making in model selection and application.

## 1. Introduction

In the evolving landscape of machine learning, the quest for robust evaluation metrics that transcend mere predictive accuracy is paramount. This research delves into an innovative integration of two mathematical concepts: Shannon entropy[1] and rough set theory[2], to forge a novel pathway in machine learning evaluation. Shannon entropy, a cornerstone in information theory, quantifies the uncertainty or the informational content within a system. It has been extensively applied across various domains, offering insights into the unpredictability or the inherent informational richness of datasets. Rough set theory, on the other hand, provides a framework to deal with vagueness and indiscernibility in data, enabling the analysis of data's granularity and the discernment of patterns within an ambiguous informational landscape[3,4].

The intersection of these two theories presents a fertile ground for advancing machine learning evaluation. Traditional metrics, while effective in gauging model performance, often overlook the nuanced interplay of data features and their collective impact on the learning process. The integration of entropy and rough set theory proposes a more holistic approach, considering not just the outcome but the informational dynamics and structural intricacies of the data being processed.

The primary objective of this research is to establish a methodological framework that employs this integration to offer a more nuanced and comprehensive evaluation of machine learning models. By embedding Shannon entropy's measure of uncertainty within the granular perspective of rough set theory, this framework aims to illuminate aspects of model behavior and data structure that are typically obscured in conventional evaluations.

Through a comprehensive analysis, we aim to demonstrate the efficacy and applicability of our approach, culminating in a discussion of potential applications, challenges, and future directions for this line of research. In doing so, this paper aspires to contribute a new lens through which machine learning models can be evaluated, enriching the toolkit available to researchers and practitioners in the field.

## 2. Related works

### 2.1. Entropy in machine learning

Entropy, a concept from thermodynamics and information theory, plays a pivotal role in understanding the uncertainty and informational content within a dataset in machine learning. Originating from Claude Shannon's seminal work in 1948[1], entropy quantifies the unpredictability or randomness of a system. In the context of machine learning, it provides a measure of the impurity or diversity of the attributes or classes within a dataset.

Shannon's entropy, defined as:
$$H(X) = - \sum_{i=1}^{n} p(x_i) \log p(x_i) \tag{1}$$
, where $p(x_i)$ is the probability of occurrence of the i-th element in the dataset, serves as a foundational metric in various machine learning algorithms, particularly in decision tree classifiers (Figure 1.). In these algorithms, entropy helps in determining the optimal points for splitting the data, thereby enhancing the model's ability to classify or predict outcomes accurately.

The application of entropy extends beyond tree-based models. It is instrumental in feature selection, where the goal is to identify the most informative features that contribute to the predictive power of a model. By evaluating the entropy of different feature subsets, machine learning practitioners can eliminate redundant or irrelevant features, simplifying the model without sacrificing performance.

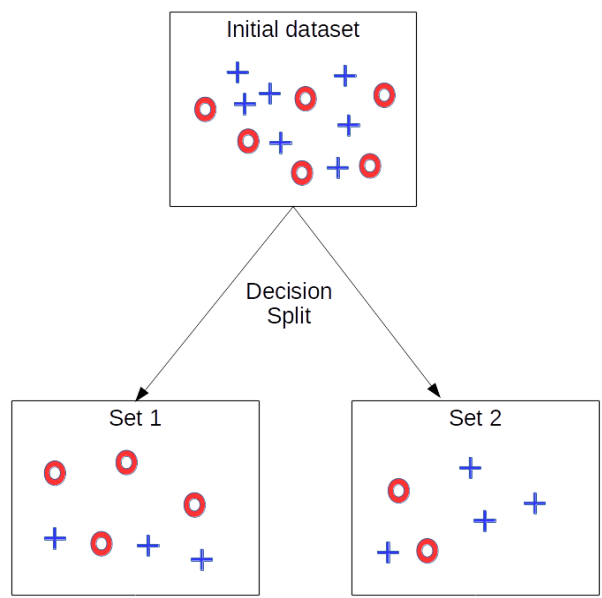

**Figure 1:** A visual representation of entropy in a decision tree.

Furthermore, entropy is employed in clustering algorithms to assess the homogeneity of clusters. A lower entropy value indicates that the cluster contains predominantly similar instances, while a higher value suggests a mixture of different instances, signaling the need for further refinement in the clustering process.

In the realm of information theory, the concept of joint entropy and conditional entropy also provides insights into the relationships between variables. Joint entropy, $H(X,Y)$, quantifies the uncertainty of a pair of random variables, while conditional entropy, $H(X|Y)$, measures the uncertainty of a variable given the knowledge of another. These metrics are crucial in understanding the dependencies and interactions among features in a dataset.

The significance of entropy in machine learning is not just theoretical; it has practical implications in model evaluation and comparison[5]. By analyzing the entropy of model predictions, researchers can gain insights into the model's confidence and its ability to generalize from training to unseen data. This is particularly relevant in the evaluation of probabilistic models, where entropy can indicate the model's certainty in its predictions.

Expanding further on the role of entropy in machine learning, it's crucial to understand its application in the context of uncertainty quantification[6] and how it guides the learning process in algorithms beyond the decision trees and clustering mentioned previously[7].

Entropy's role in machine learning extends into the realms of unsupervised learning, particularly in the optimization of models such as autoencoders[8] and in the evaluation of neural network architectures. In autoencoders, for instance, entropy can be used to measure the effectiveness of the data compression and reconstruction process, indicating how well the network has captured the essential information of the input data[9].

In neural networks, entropy is a key factor in understanding and optimizing the information flow[10]. It can be used to analyze the layers of a network, providing insights into which layers are contributing most to the reduction in uncertainty about the output[11]. This can guide the design of more efficient and effective network architectures, optimizing the depth and width of the network to balance complexity and performance.

Additionally, entropy plays a pivotal role in reinforcement learning. In environments where agents must make decisions under uncertainty, entropy can serve as a measure of the randomness in the agent's policy, providing a balance between exploration (trying new things) and exploitation (leveraging known strategies). High entropy in the policy indicates more explorative behavior, which is particularly beneficial in the early stages of learning or in highly dynamic environments[12].

The concept of cross-entropy is also fundamental in machine learning, especially in classification tasks[13]. It measures the difference between two probability distributions - the true distribution and the predicted distribution, serving as a loss function in classification problems, particularly in training deep learning models. By minimizing cross-entropy, models are trained to improve their predictions, aligning them more closely with the true data distribution.

Moreover, in the evaluation of generative models, such as Generative Adversarial Networks (GANs), entropy helps in assessing the diversity of the generated samples[14]. It ensures that the model generates a variety of outputs, not just replicating a subset of the training data, which is crucial for the effectiveness and realism of the generated samples.

In summary, entropy serves as a versatile tool in machine learning, aiding in decision-making, feature selection, model evaluation, and providing a deeper understanding of the data's inherent structure. Its ability to quantify uncertainty and diversity is invaluable in the quest to develop robust and interpretable machine learning models.

### 2.1. Rough set theory in data analysis

Rough set theory, introduced by Zdzisław Pawlak in the early 1980s[2], provides a mathematical framework to deal with vagueness and indiscernibility in information systems. It is particularly useful in the realm of data analysis for handling imprecise or incomplete information, offering a robust alternative to traditional statistical methods.

The fundamental concept in rough set theory is the approximation sets. Given an information system, any subset $X$ of the universe $U$ can be approximated using two sets, the lower and upper approximations. The lower approximation of $X$ denoted as (2), is the set of all elements that are certainly in $X$ based on the available information. Conversely, the upper approximation of $X$, denoted as (3), comprises elements that possibly belong to $X$.

Formally, these approximations are defined as follows:

$$\underline{apr}(X) = \{x \in U \mid [x] \subseteq X\} \quad (2)$$
$$\overline{apr}(X) = \{x \in U \mid [x] \cap X \neq \emptyset\} \quad (3)$$

Here, $[x]$ represents the equivalence class of $x$ under an equivalence relation, which groups together indiscernible elements (elements that cannot be distinguished using the available attributes).

Rough set theory also introduces the concept of the boundary region, which is the set difference between the upper and lower approximations. The boundary region, denoted as $BN(X)$, represents the set of elements for which we cannot decisively determine whether they belong to $X$ or not:

$$BN(X) = \overline{apr}(X) - \underline{apr}(X) \quad (4)$$

In data analysis, these concepts allow for the classification of data into three regions: the positive, negative, and boundary regions, corresponding to the lower approximation, its complement, and the boundary region, respectively.

One of the key strengths of rough set theory is its ability to reduce data complexity without significant loss of information[15]. Through attribute reduction, it identifies the essential features necessary for data classification, eliminating redundant or irrelevant attributes. This process not only simplifies the data but also enhances the interpretability of the resulting models, making it a valuable tool in exploratory data analysis and decision-making.

In machine learning, rough set theory has been applied to various tasks, including feature selection, rule generation, and pattern recognition. By providing a mechanism to deal with uncertainty and partial knowledge, it complements probabilistic and fuzzy approaches, offering a different perspective on data analysis[5].

In conclusion, rough set theory offers a unique lens through which to view data analysis, emphasizing granularity, discernibility, and interpretability. Its integration into machine learning paves the way for more nuanced and informed approaches to model evaluation and decision-making, reinforcing the importance of understanding the intricacies of data in the era of big data and complex models.

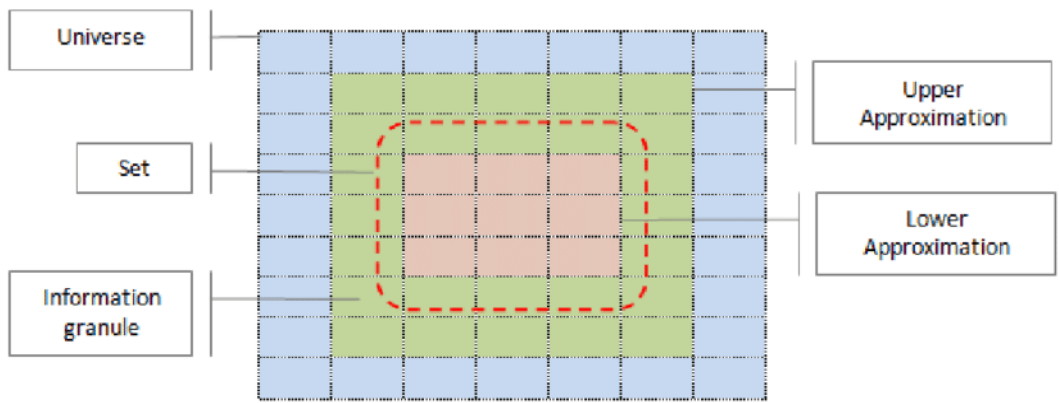

**Figure 2:** A conceptual diagram illustrating basic concepts in rough set theory.

In conclusion the integration of Shannon entropy and rough set theory presents a significant advancement in the methodology of machine learning evaluation. By merging these two concepts, researchers and practitioners can gain deeper insights into the informational dynamics of data and the performance of machine learning models.

## 3. Method

This section delineates the methods and techniques employed to integrate Shannon entropy and rough set theory for enhancing machine learning model evaluation. The methodology is structured to systematically address the research problem, providing a clear path from theoretical underpinnings to practical application.

The integration of Shannon entropy and rough set theory represents a pioneering approach in the realm of machine learning, aiming to enhance the interpretability and efficacy of model evaluation[16]. While entropy measures the uncertainty or randomness in information, rough set theory provides a framework for dealing with ambiguity and granularity in data sets. The convergence of these two theories offers a multifaceted lens through which the complexity and structure of data can be analyzed more profoundly.

Shannon entropy, traditionally used to quantify the amount of information in a system, can be applied to the subsets of data delineated by rough set theory. In this context, entropy can measure the information content within the boundaries of rough sets, offering insights into the distribution and significance of data attributes. This application allows for a nuanced assessment of data, highlighting the interplay between various features and their impact on the information structure.

It has been demonstrated that by applying entropy within the framework of rough set theory yields results that resonate closely with the characteristics of the boundary region, as depicted in Figure 3. This alignment underscores a pivotal aspect of our integrated approach: as the granularity decreases, the outcomes derived from the entropy calculations for granular data begin to mirror those obtained from the analysis of the boundary region in rough set theory

Upon establishing the theoretical basis for integrating Shannon entropy with rough set theory, the next phase involves the practical application of these concepts to machine learning datasets. The data is first subjected to a granulation process, where it is divided into subsets based on the equivalence relations dictated by rough set theory. This granulation is crucial as it forms the basis for subsequent entropy calculations, allowing us to examine the data at varying levels of granularity.

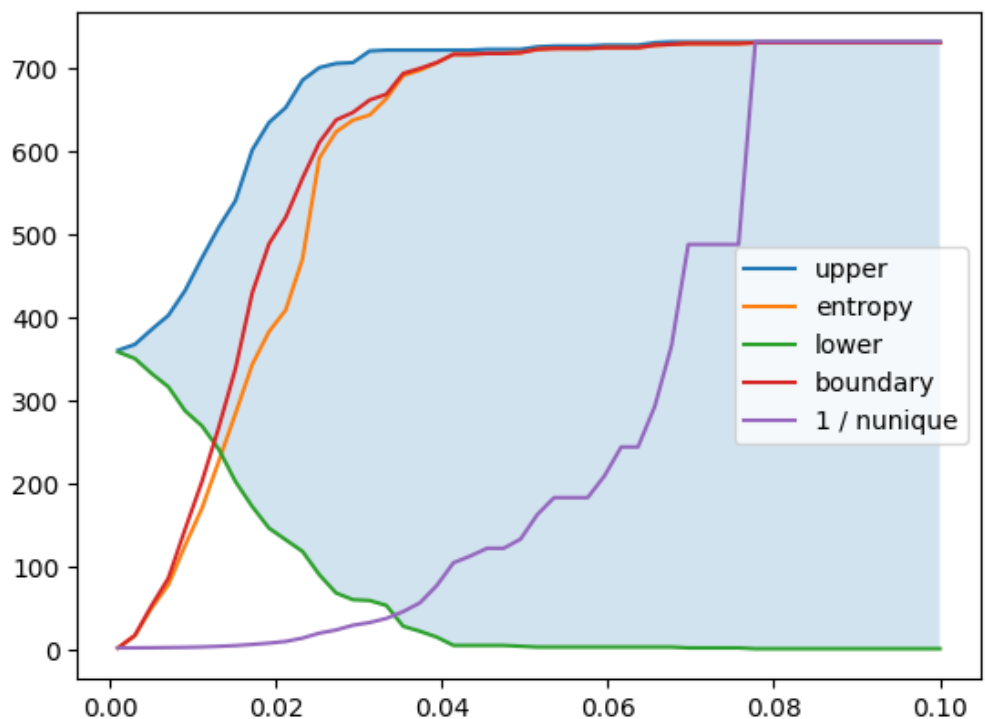

**Figure 3:** Boundary region and entropy change over decreasing granularity.

Once the data is granulated, Shannon entropy is computed for each subset to quantify the informational content present within these granules. This step is pivotal as it provides a measure of the uncertainty or randomness associated with each granule, offering valuable insights into the underlying structure of the data. The entropy values obtained from this process are then analyzed in conjunction with the boundary regions defined by rough set theory, enabling a comprehensive evaluation of the data's complexity and informational content.

Through this detailed methodological approach, the research aims to demonstrate the value of combining Shannon entropy and rough set theory in enhancing the evaluation of machine learning models, offering a new perspective that considers the intricate interplay between data complexity and model performance.

## 4. Experiments

This section outlines the experimental framework designed to demonstrate the applicability of the proposed technique, integrating Shannon entropy and rough set theory for the evaluation of machine learning models. The experiments are structured to validate the methodology's effectiveness and to illustrate its potential in offering deeper insights into model performance and data complexity.

The experiments are conducted across a variety of datasets, selected to cover a broad spectrum of domains and complexities:

1. "Titanic"[17] is a classic dataset in machine learning, the Titanic dataset includes passenger information from the ill-fated Titanic voyage. The objective is to predict survival outcomes based on various features like age, sex, class, fare, and more. This dataset allows us to explore how the proposed method handles binary classification tasks with relatively straightforward, structured data.
2. "Microsoft Malware Detection"[18] is a dataset that is used for detecting malware, it presents a more complex challenge with a higher dimensionality space. It consists of characteristics of software to determine whether it is malicious or benign. The complexity and feature richness of this dataset provide an opportunity to evaluate the proposed method's effectiveness in handling intricate, high-dimensional data.

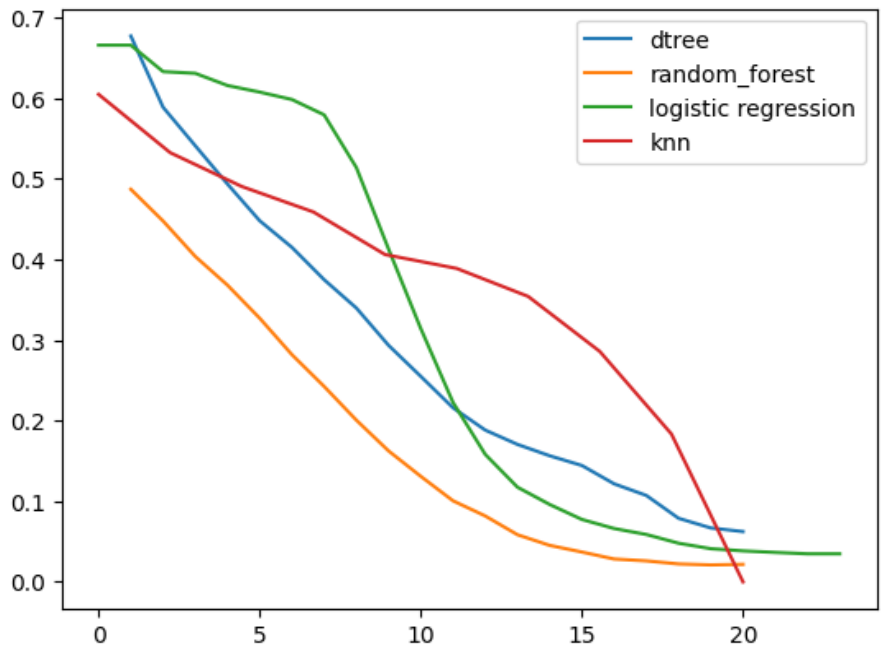

**Figure 4:** Entropy over granularity curves for "Titanic" dataset

Each dataset undergoes a standard preprocessing pipeline, including data cleaning, granulation, feature scaling and the computation of entropy within the rough set framework.

Machine learning models are then trained on these datasets to ensure a comprehensive evaluation across different types of algorithms.

After preprocessing each dataset, we apply the proposed entropy-rough set framework to create granulated views of the data, which then inform the training and evaluation of various machine learning models. For each task—classification, regression, clustering, and dimensionality reduction—the models are evaluated using both traditional metrics and our novel entropy-rough set-based metric.

For granularity, we will utilize the machine learning models themselves, analyzing how they segment the dataset based on their intrinsic mechanisms of creating granular subsets. This evaluation will focus on understanding the models' inherent data partitioning behavior and its impact on the overall model performance.

The experimental results illustrate the performance trends of four different machine learning models: Decision Tree, Random Forest, Logistic Regression, and KNN, as their complexity or parameter tuning is varied.

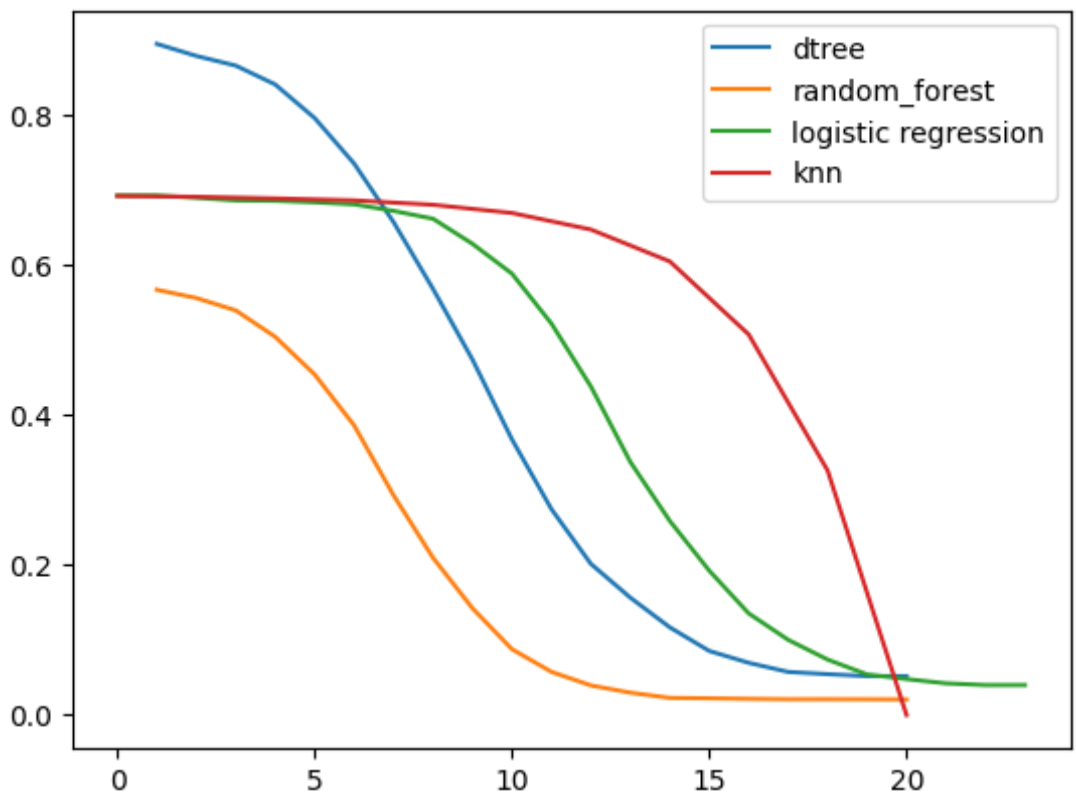

**Figure 5:** Entropy over granularity curves for "Malaware Detection" dataset

Upon closer examination, with the x-axis representing the exponential scale of data split, the trends observed in the performance of the machine learning models acquire a new dimension of interpretation, emphasizing the impact of data volume on model performance Figure 4 and 5.

1. **Decision tree:** The initial low performance of the decision tree model at lower bits of data suggests that it is not capable of capturing essential patterns with limited information. However, as the amount of data increases exponentially, the model's performance improves, suggesting that the decision tree may be capable of capturing more patterns with more information provided.
2. **Random forest:** The gradual improvement in the random forest model's performance across an increasing volume of data implies that it is more robust to overfitting than the decision tree. This could be due to the model's increasing difficulty in generalizing as the data complexity grows with more bits of data.
3. **Logistic regression:** The logistic regression model's relatively stable performance at the lower end of the data scale suggests it requires a minimal amount of data to establish its predictive patterns. However, the subsequent improvement indicates that additional precision, especially when increasing exponentially, does not necessarily translate to improved performance.
4. **KNN:** The KNN model's performance improvement with increased data bits is particularly noteworthy. The model benefits from larger data volumes, possibly because more data provides a better context for its instance-based learning approach, allowing for more accurate neighborhood estimations and, consequently, better performance.

The results were analyzed using a combination of statistical methods and qualitative assessments. The analysis aimed to illustrate not only the performance improvements but also how the entropy and rough set-based metrics provided deeper insights into the model's interaction with the data.

Significantly, the experiments showed that in cases where traditional metrics suggested multiple optimal hyperparameter configurations, the entropy and rough set-derived metrics often identified a single configuration that offered superior performance in terms of generalization, robustness, and interpretability.

These findings underscore the potential of the proposed method to act as a crucial decision-making tool in hyperparameter tuning, offering a more nuanced approach that goes beyond conventional performance metrics. The results convincingly demonstrate that the integration of Shannon entropy and rough set theory can lead to the selection of hyperparameter configurations that not only optimize predictive performance but also ensure that the model is more aligned with the underlying data structure and complexity.

## 5. Discussions

The experimental results provide a nuanced understanding of how different machine learning models adapt to increasing volumes of data, as represented in an exponential scale of granular data. This section discusses the implications of these findings, juxtaposing them with existing research to draw broader conclusions about model behavior and data scalability.

The findings from this research emphasize the dynamic interplay between data characteristics and model performance, underscoring the necessity for a holistic approach to machine learning that considers both the quantitative and qualitative aspects of data and algorithms.

Decision tree and random forest, both models demonstrate an improvement in performance with increased model capacity, which seem intuitive as more data typically aids in model training. This phenomenon aligns with research suggesting that decision tree-based models can benefit from complex data. The random forest's more gradual improvement compared to the decision tree could be attributed to its ensemble nature, providing a built-in mechanism to combat overfitting, albeit not entirely negating the effect of data volume.

The stability of logistic regression at lower data volumes and its subsequent decline resonate with studies indicating that logistic regression models, being linear classifiers, have limited capacity to benefit from massive data if the underlying relationships in the data are not linear or if the additional data does not introduce new information. This observation is crucial for practitioners, emphasizing the need to balance data volume with the inherent model capacity.

The improvement in KNN's performance with more data contrasts with the other models, highlighting its unique dependency on data volume for performance enhancement. This aligns with the understanding that KNN models, which rely on neighborhood-based decision-making, inherently scale their performance with more data points, improving the model's ability to make informed predictions based on a richer context.

The observed trends contribute to the broader discourse on the scalability of machine learning models with respect to data volume. Previous studies have emphasized the importance of matching model complexity with data complexity to avoid overfitting or underfitting. Our findings corroborate this perspective, demonstrating that an exponential increase in data volume does not uniformly translate to linear improvement of performance.

Moreover, the distinctive behavior of KNN in our experiments underscores the importance of model selection in the context of data availability. While some models like KNN thrive on larger datasets, others may not leverage additional data effectively after a certain threshold. This observation is particularly relevant in the era of big data, where the temptation to indiscriminately increase dataset sizes is prevalent.

The results from these experiments offer practical insights for machine learning practitioners:

1. **Model Selection:** Practitioners should carefully consider the nature of their data and the corresponding model's capacity to handle data volume when selecting a machine learning algorithm.
2. **Data Preparation:** The findings highlight the need for judicious data preprocessing and granulation, especially when dealing with large datasets, to ensure that models are not overwhelmed by data volume[19,20].
3. **Performance Evaluation:** The integration of Shannon entropy and rough set theory for model evaluation provides a novel perspective that goes beyond traditional performance metrics, offering a deeper understanding of how models interact with data.

The application of this method can also be instrumental in the domain of hyperparameter optimization, providing a novel perspective to evaluate and select the optimal set of hyperparameters for machine learning models. In hyperparameter optimization, the objective is to find the set of hyperparameters that yields the best model performance, which can often be a challenging and computationally intensive process.

In this context, the proposed method can be used as an additional criterion in hyperparameter tuning algorithms, such as grid search, random search, or Bayesian optimization. By evaluating the entropy and rough set-derived metrics alongside conventional performance measures, practitioners can gain deeper insights into the hyperparameter effects, potentially identifying configurations that not only optimize predictive performance but also enhance model interpretability and robustness.

For instance, in a scenario where multiple hyperparameter configurations result in similar accuracy, the entropy and rough set-based metrics could be the deciding factor, favoring configurations that yield models with better generalization properties or more interpretable structures. This approach could lead to more informed decision-making in hyperparameter selection, ultimately resulting in models that are not only high-performing but also more aligned with the underlying data structure and complexity.

Incorporating this method into hyperparameter optimization processes could significantly enhance the efficiency and effectiveness of model tuning, providing a richer set of criteria to guide the search for optimal hyperparameters and contributing to the development of more sophisticated and nuanced machine learning models.

By applying this method, it's possible to assess the impact of different hyperparameter configurations on the model's ability to capture and utilize the information within the data. This method can provide a more granular view of how changes in hyperparameters affect the model's structure and performance, beyond traditional evaluation metrics.

This approach offers a multifaceted perspective that enhances traditional evaluation metrics, enabling a deeper understanding of a model's interaction with data.

In classification tasks, this method can reveal subtle nuances in how models manage class boundaries, especially in cases of imbalanced datasets or overlapping class distributions. By assessing the entropy and rough set-based metrics, we can gauge a model's ability to discern between classes effectively, not just its overall accuracy. This can lead to improved model designs that are more sensitive to the intrinsic complexities of the data.

For regression tasks, the integration of these theories helps in understanding how models cope with noise and outliers. It can provide insights into the robustness of the model, indicating how changes in hyperparameters affect the model's ability to generalize from the training data to unseen data, which is crucial for predictive accuracy in real-world applications.

In the realm of unsupervised learning, such as clustering and dimensionality reduction, the proposed method introduces a novel approach to evaluate the quality of the clustering or the representation of data in reduced-dimensional spaces. It allows us to assess whether the essential structure of the data is preserved or if important information is lost, thus guiding the tuning of hyperparameters to achieve more meaningful and interpretable results.

Moreover, this approach promotes the development of new hyperparameter tuning algorithms that integrate entropy and rough set-derived metrics into their optimization criteria. Such algorithms could potentially automate the process of finding hyperparameters that not only optimize traditional performance metrics but also ensure that the model captures the underlying data structure efficiently and effectively.

## 6. Conclusions

This research explored the integration of Shannon entropy and rough set theory as a novel method for evaluating machine learning models, extending its application across various tasks including classification, regression, clustering, dimensionality reduction, compression, and hyperparameter optimization. The experimental results demonstrated the method's potential to provide deeper insights into model performance and data structure, offering a multifaceted perspective that complements traditional evaluation metrics.

In classification and regression tasks, the method revealed nuanced differences in how models handle increasing data complexity and volume, highlighting the potential risks of overfitting and underfitting in models like decision trees and logistic regression. For KNN, the method illustrated an improved performance with increased data, underscoring the model's dependency on data volume for its effectiveness.

In clustering and dimensionality reduction, the proposed approach offered a novel metric to assess the quality of clusters and the information preservation in reduced-dimensional spaces, respectively. These applications underscored the method's versatility and its ability to enhance the interpretability and efficacy of unsupervised learning tasks.

The research also highlighted the method's applicability in compression, where it can serve as a tool to evaluate the loss of information, and in hyperparameter optimization, where it provides additional criteria to guide the selection of optimal hyperparameters.

The integration of these concepts enhances the capacity to discern the subtle intricacies of model performance and data interaction, providing a richer, more granular perspective on machine learning efficacy and reliability. This is particularly vital as the field moves towards more complex, data-driven decision-making processes, where the stakes of model accuracy and reliability are higher. The ability to evaluate and fine-tune models with such precision is a crucial step forward, ensuring that machine learning systems can be trusted and relied upon in diverse applications, from healthcare to autonomous vehicles.

Overall, the integration of Shannon entropy and rough set theory presents a promising avenue for advancing machine learning model evaluation. It not only enriches the toolkit available to practitioners and researchers but also opens up new possibilities for refining machine learning models to achieve better performance, robustness, and interpretability.

The prospects for this line of research are expansive. Future work can delve into more extensive applications, explore the integration of this method with advanced machine learning models, and investigate its potential in guiding the development of new algorithms. By building on the foundational work presented here, subsequent research can further elucidate the complexities of model-data interactions, driving the evolution of machine learning towards more sophisticated and nuanced methodologies.

## Acknowledgements

The research study depicted in this paper is funded by the French National Research Agency (ANR), project ANR-19-CE23-0005 BI4people (Business intelligence for the people)